\documentclass[runningheads]{llncs}
\usepackage{graphicx}
\usepackage{amsmath}
\usepackage{adjustbox}
\usepackage{amssymb}
\usepackage{arydshln}
\usepackage{hyperref}
\hypersetup{
    hidelinks,
    colorlinks=true,
    citecolor=black,
    linkcolor=black,
    urlcolor=blue,
    pdftitle={Title},
    pdfauthor={Author}
}

\begin{document}

\title{CDER: Collaborative Evidence Retrieval for Document-level Relation Extraction}

\titlerunning{CDER: Collaborative Evidence Retrieval for DocRE}

% \author{Khai Phan Tran\thanks{Corresponding Author.}\orcidID{0000-0001-9870-4185} \and Xue Li\orcidID{0000-0002-4515-6792}}
\author{Khai Phan Tran\thanks{Corresponding Author.} \and Xue Li}

\authorrunning{K. P. Tran et al.}

\institute{
    School of Electrical Engineering and Computer Science, 
    \\ The University of Queensland, Brisbane, Australia 
    \\ \email{phankhai.tran@uq.edu.au}, \ \email{xue.li@eecs.uq.edu.au}
}

\maketitle

\begin{abstract}
    Document-level Relation Extraction (DocRE) involves identifying relations between entities across multiple sentences in a document. Evidence sentences, crucial for precise entity pair relationships identification, enhance focus on essential text segments, improving DocRE performance. However, existing evidence retrieval systems often overlook the collaborative nature among semantically similar entity pairs in the same document, hindering the effectiveness of the evidence retrieval task. To address this, we propose a novel evidence retrieval framework, namely CDER. CDER employs an attentional graph-based architecture to capture collaborative patterns and incorporates a dynamic sub-structure for additional robustness in evidence retrieval. Experimental results on the benchmark DocRE dataset show that CDER not only excels in the evidence retrieval task but also enhances overall performance of existing DocRE system\footnote{Our code is available at: \url{https://github.com/khaitran22/CDER}
}.

\end{abstract}

\section{Introduction}
\label{intro}
Document-level Relation Extraction (DocRE), drawing increased research interest recently, leverages contextual information across multiple sentences to identify relations between entity pairs \cite{yao2019docred}. Therefore, DocRE presents a greater challenge, requiring comprehensive reading and reasoning over multiple sentences to prioritize relevant contexts, compared to sentence-level relation extraction. To tackle DocRE, solutions include using pre-trained language models as document encoders \cite{tan2022document,zhou2021document} or constructing heterogeneous graphs to explicitly model information interactions between entities \cite{wang2020global,zeng2020double}. To enhance DocRE models' resilience against noisy information, evidence retrieval (ER) in DocRE is another direction. ER involves retrieving a minimal subset of sentences, termed ``\textit{evidence sentences}", essential for accurately predicting relations between entity pairs \cite{yao2019docred}. Existing studies \cite{lu2023anaphor,ma2023dreeam,xiao2021sais,xie2022eider} have been underscoring the significance of ER in improving DocRE performance.

\begin{figure}[t]
    \centering
    \includegraphics[width=\linewidth]{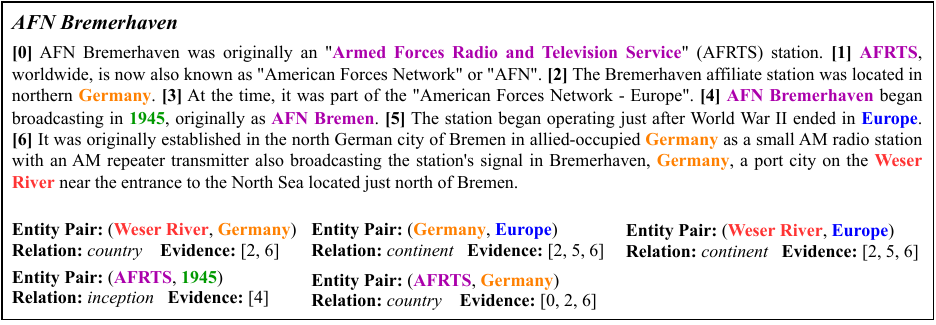}
    \caption{An example document from the DocRED dataset \cite{yao2019docred}. Entity pairs are given with corresponding supporting evidence sentences (\textit{\textit{i.e.},} \textbf{Evidence}) and relations.}
    \label{fig:example_docred}
\end{figure}

Recent ER models, like \cite{ma2023dreeam,xie2022eider}, have shown promising progress in identifying precise evidence sentences for entity pairs, but they often retrieve evidence for each entity pair separately, potentially leading to sub-optimal ER performance. Notably, entity pairs within the same document exhibit a collaborative tendency where evidence sentence sets overlap significantly for pairs sharing common entities and semantically correlated relations. For instance, in Figure \ref{fig:example_docred}, entity pairs (\textit{Weser River}, \textit{Europe}), (\textit{Weser River}, \textit{Germany}), and (\textit{Germany}, \textit{Europe}) exhibit overlapping evidence sentence sets due to shared entities and similar relations. Consequently, it is plausible to hypothesize that leveraging this collaborative nature can enhance ER outcomes and subsequently improve the overall DocRE performance.

Motivated by this, we introduce a framework called \textbf{C}ollaborative \textbf{D}ocument-level \textbf{E}vidence \textbf{R}etrieval (CDER) to explicitly model aforementioned collaborative nature. CDER transforms the document into a bipartite graph with entity pair and sentence nodes, partitioning it into three sub-graphs: (i) Entity Pair sub-graph models collaboration, (ii) Entity Pair-Sentence sub-graph captures semantic relationships, and (iii) Sentence sub-graph aids co-reference reasoning. The Entity Pair sub-graph is equipped with a novel dynamic structure for better flexibility in modeling collaboration. To address noisy information, we introduce Entity Pair-aware Graph Attention Networks (GATs) inspired by Graph Attention Networks (GATs) \cite{velivckovic2018graph}, tailored to our framework. Experiments on the DocRED dataset \cite{yao2019docred} demonstrate the effectiveness of CDER framework, with the generated evidence sentences enhancing baseline DocRE systems.

In summary, our main contributions are:
\begin{itemize}
    \item We propose and explore the concept of collaborative nature among semantically similar entity pairs in the ER task.
    \item We devise a novel attentional graph-based framework with a dynamic structure, explicitly designed to harness collaborative nature information.
    \item Experimental results validate the efficacy of our approach, showcasing its superiority over previous state-of-the-art ER systems.
\end{itemize}

\section{Related Works}
\subsubsection{Evidence Retrieval in DocRE.}
Early DocRE approaches \cite{tan2022document,zhang2021document,zhou2021document} utilized entire documents as context input, despite the need for only a few crucial evidence sentences for precise entity pair relation extraction (RE).

Huang et al. \cite{huang2021three} identified that over 95\% of entity pairs require no more than 3 sentences as supporting evidence and introduced heuristic rules for evidence identification. Huang et al. \cite{huang2021entity} later proposed a joint training framework for RE and ER, but did not leverage evidence results for RE improvement, and their ER model had prolonged training times. In response, Xie et al. \cite{xie2022eider} introduced a lightweight ER model named EIDER and a fusion strategy to enhance RE using evidence results. Additionally, Ma et al. \cite{ma2023dreeam} used attention modules from DocRE systems to detect evidence, further reducing memory usage compared to EIDER.

Nevertheless, existing models tend to retrieve evidence sentences for each entity pair independently, often neglecting the collaborative nature among entity pairs. In contrast, our proposed method directly exploits this collaboration, with the goal of achieving superior ER performance compared to previous approaches.

\section{Methodology}
\subsection{Problem Formulation}
In a document $D$ with a set of $n$ tokens $\mathcal{T}_D= \{t_i\}_{i=1}^n$ and sets of sentences $\mathcal{S}_D = \{s_i\}_{i=1}^{|\mathcal{S}_D|}$ and entities $\mathcal{E}_D=\{e_i\}_{i=1}^{|\mathcal{E}_D|}$, each entity $e_i$ has a mention set $\mathcal{M}_{e_i}=\{m^i_j\}_{j=1}^{N_{e_i}}$, where $m^i_j$ is the $j^{\text{th}}$ mention of $e_i$, and $N_{e_i}$ is the mention count. In the DocRE task, the objective is to identify a relation set $\mathcal{R}_{h,t}$ from a pre-defined relation sets $\mathcal{R}$ between the entity pair $(e_h, e_t)$. Here, $e_h$ and $e_t$ are the head and tail entities, and $\mathcal{R}_{h,t} \subset \mathcal{R}$ or $\mathcal{R}_{h,t} = {NA}$ denotes ``no relation". For a valid relation $r \in \mathcal{R}$ between $(e_h, e_t)$, the ER task aims to extract a set of evidence sentences $\mathcal{S}_{h,r,t} \subseteq \mathcal{S}_D$ needed to predict relation $r$.

\subsection{Encoder Module}
\subsubsection{Document Encoding}
For the document $D$, we insert a special token ``\textbf{*}'' before and behind each mention $m_j^i$ of $e_i$ as an entity marker \cite{zhang2017position}. We feed document $D$ into a pre-trained language model (PLM) with a dimension of $d$ to obtain token embeddings $H \in \mathbb{R}^{n \times d}$ from the last three transformer layers of the PLM:
\begin{equation}
    H, A = PLM([t_1, t_2, ..., t_n])
\end{equation}

\subsubsection{Entity Embedding}
\vspace{-3mm}
The embedding $\mathbf{e}_i$ for each entity $e_i$ is computed by applying the \texttt{logsumexp} operation over the representations of the starting token ``\textbf{*}" (denoted as $\mathbf{m}_j^i$) for each mention $m_j^i$ of $e_i$.
\begin{equation}
    \mathbf{e}_i = \text{log} \sum_{i=1}^{N_{e_i}}\text{exp}\left(\mathbf{m}_j^i\right)
\end{equation}

Subsequently, the representation $\mathbf{p}$ of an entity pair $p = (e_h, e_t)$ is defined as:
\begin{equation}
    \mathbf{p} = \text{tanh}\left(\mathbf{W}_p\left[\mathbf{e}_h; \mathbf{e}_t; \mathbf{c}_{h, t}\right] + \mathbf{b}_p\right)
\end{equation}
where $\mathbf{W}_p$ and $\mathbf{b}_p$ are trainable parameters;  $[ \ ; \ ]$ is the concatenation; $\mathbf{c}_{h, t}$ is the localized context embedding of $(e_h, e_t)$ proposed in \cite{zhou2021document}.

\begin{figure*}[t]
    \centering
    \includegraphics[width=0.86\textwidth]{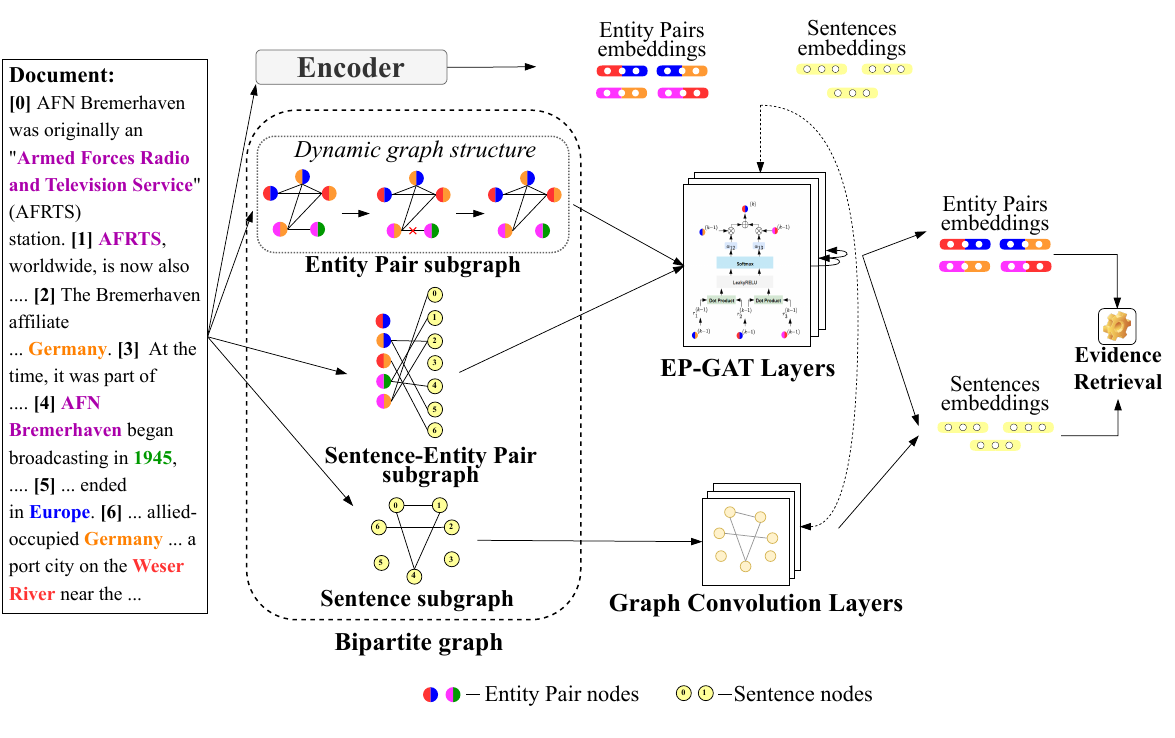}
    \caption{\textbf{Overall framework of CDER}: We obtain entity pairs and sentence embeddings and construct a bipartite graph from the document. Then, the graph is divided into Entity Pair ($G_{PP}$), Sentence-Entity Pair ($G_{PS}$), and Sentence ($G_{SS}$) sub-graphs. We enhance the embeddings of entity pairs and sentences using our proposed EP-GAT layers on $G_{PP}$ and $G_{PS}$, and GCNs \cite{kipf2016semi} on $G_{SS}$. The final refined embeddings are employed for ER tasks.}
    \label{fig:overall-cder}
    \vspace{-2mm}
\end{figure*}

\subsubsection{Sentence Embedding}
We apply the mean operator on representations $\mathbf{h}^i_j$ of all $N_{s_i}$ tokens $h^i_j$ within the sentence $s_i$ to obtain its representation $\mathbf{s}_i$:
\begin{equation}
    \mathbf{s}_i = \frac{1}{N_{s_i}}\sum_{j=1}^{N_{s_i}} \mathbf{h}^i_j
\end{equation}

\subsection{Document-level Bipartite Graph}
\label{graph}
We transform the document into a bipartite graph with entity pair and sentence node types. This graph is further decomposed into three sub-graphs: (i) Entity Pair, (ii) Entity Pair-Sentence, and (iii) Sentence. Figure \ref{fig:overall-cder} depicts our proposed framework.

\subsubsection{Entity Pair Sub-graph ($G_{PP}$)}
The sub-graph, denoted as $G_{PP} = (V_{PP}, E_{PP})$, is created to model the collaborative nature between semantically similar entity pairs, as introduced in Section \ref{intro}. It consists of a set of entity pair nodes $V_{PP}$ from document $D$ and a set of p2p (entity pair-to-entity pair) edges $E_{PP}$. Initially, two nodes $p_1, p_2 \in V_{PP}$ are connected in $G_{PP}$ if their corresponding pairs share a common entity. An example is (\textit{Weser River}, \textit{Germany}) and (\textit{Germany}, \textit{Europe}) from Figure \ref{fig:example_docred}.

To further enhance the collaborative nature, we introduce a dynamic structure for the $G_{PP}$. Intuitively, if initially connected entity pairs' relations lack significant semantic correlation, their supporting evidence sets become disjoint. For instance, evidence sets of  (\textit{AFRTS}, \textit{1945}) and (\textit{AFRTS}, \textit{Germany}) from Figure \ref{fig:example_docred} do not overlap due to their semantically irrelevant relations. Consequently, aggregating information between such pairs can introduce noise and degrade ER performance during inference.

To accomplish the objective, we compute the relevance of any two pairs $p_i, p_j \in V_{PP}$, denoted as $sim(p_i, p_j)$. The relevance score $sim(p_i, p_j)$ is defined as:
\begin{equation}
    sim(p_i, p_j) = \mathbf{r}_i \cdot \mathbf{r}_j
\end{equation}
where $\mathbf{r}_i, \mathbf{r}_j$ are relational information of $p_i, p_j \in V_{PP}$ computed by TransE \cite{bordes2013translating} as the relations of entity pairs is absent during testing stage and $(\cdot)$ is dot product operator. Subsequently, we either remove edges between $p_i, p_j$ if they are connected and $sim(p_i, p_j) < \theta$ or add edges between $p_i, p_j$ if they are disconnected and $sim(p_i, p_j) \geq \theta$ during runtime.

During training, $V_{PP}$ only contain nodes from positive entity pairs (\textit{i.e.,} those with a valid relation $r$). However, because positive entity pairs are unknown during inference, $V_{PP}$ will contain nodes from all possible entity pair permutations in a document $D$. Therefore, $V_{PP}$ contains $|\mathcal{E}_D| \times \left(|\mathcal{E}_D|-1\right)$ entity pair nodes during inference.

\subsubsection{Entity Pair-Sentence Sub-graph ($G_{PS}$)}
To encapsulate the semantic relationships between sentences and entity pairs in document $D$, we construct the Entity Pair-Sentence sub-graph. The sub-graph, denoted as $G_{PS} = (V_{PS}, E_{PS})$ where $V_{PS} = V_{PP} \cup V_{SS}$, encompasses a node set $V_{PS}$ comprising entity pair nodes $V_{PP}$ and sentence nodes $V_{SS}$, and a collection of p2s (entity pair-to-sentence) and s2p (sentence-to-entity pair) edges, designated as $E_{PS}$. In $G_{PS}$, a node $p_i \in V_{PP}$ connects to a node $s_j \in V_{SS}$ if $s_j$ references at least one entity within $p_i$. An example is sentence [2] and (\textit{AFRTS}, \textit{Germany}) in Figure \ref{fig:example_docred}. Notably, the mentions and positions of entities are predetermined by datasets.

\subsubsection{Sentence Sub-graph ($G_{SS}$)}
The Sentence sub-graph $G_{SS}$ is introduced to model co-reference reasoning among sentences in document $D$. The sub-graph, denoted as $G_{SS} = (V_{SS}, E_{SS})$, comprises sentence nodes $V_{SS}$ from dataset $D$ and s2s (sentence-to-sentence) edges $E_{SS}$. In $G_{SS}$, an edge connects two sentence nodes, $s_i, s_j \in V_{SS}$, when they reference the same entity. An example is sentence [0] and sentence [1] in Figure \ref{fig:example_docred}. This information about sentence-entity associations is available in datasets in advance.

\subsection{Entity Pair-aware GATs}
\label{ep-gat}
This section presents our Entity Pair-aware GAT (EP-GAT) layer, inspired by the Graph Attention Networks (GATs) layer \cite{velivckovic2018graph}. EP-GAT is applied to both $G_{PS}$ and $G_{PP}$. We provide a detailed design of the EP-GAT layer on $G_{PS}$ as an illustrative example, and the process for $G_{PP}$ is analogous.

\subsubsection{EP-GAT Motivation} 
 Intuitively, not all sentences heuristically connected to a pair $p_i$ during construction of $G_{PS}$ serve as evidence sentences. For instance, (\textit{AFRTS}, \textit{1945}) is connected to sentences [0], [1], and [4], but only sentence [4] is evidence from Figure \ref{fig:example_docred}. Therefore, uniformly aggregating information from all connected sentences into $p_i$ can introduce noise and degrade performance.

In $G_{PP}$, following the employment of the dynamic structure to remove edges linking irrelevant entity pairs, among the remaining connected neighbor pairs of the target pair, we prioritize the influence of more significant entity pairs on shaping the representation. Thus, we incorporate the EP-GAT layer into $G_{PP}$ to fulfill this objective.

In $G_{SS}$, there are no significant differences among neighbours of a sentence as the sub-graph primarily emphasizes on modelling co-reference reasoning. Consequently, we utilize Graph Convolutional Networks layers \cite{kipf2016semi} on $G_{SS}$.

\subsubsection{EP-GAT Architecture}
\paragraph{In $G_{PS}$.} In the EP-GAT layer of $G_{PS}$, the initial step involves calculating attention weights $\mathbf{\alpha}^{(PS)}_{ij}$ between an entity pair $p_i$ and a sentence $s_j$. This computation is carried out using the localized context embedding $\mathbf{c}_{p_i}$ \cite{zhou2021document} of the pair and the sentence embedding $\mathbf{s}_j$ as inputs. Following GATs, we incorporate the self-attention mechanism before expanding to the multi-head attention mechanism \cite{vaswani2017attention} to enhance the stability of the learning process. The entire process of computing attention weight $\mathbf{\alpha}^{(PS)}_{ij}$ between $p_i$ and $s_j$ as follows:
\begin{equation}
    \label{attn-mecha}
    \footnotesize
    \begin{gathered}
        \mathbf{h}^{(PS)}_{c_{p_i}} = \mathbf{W}_{c_2} \left( \mathbf{W}_{c_1} \mathbf{c}_{p_i} \right), \\ 
        \mathbf{h}^{(PS)}_{s_j} = \mathbf{W}_{s_2} \left( \mathbf{W}_{s_1} \mathbf{s}_j \right) \\
         {\mathbf{\alpha}^{(PS)}_{ij} = \frac{\text{exp} \left( \text{LeakyRELU} \left( \mathbf{h}^{(PS)}_{c_{p_i}} \cdot \mathbf{h}^{(PS)}_{s_j} \right) \right) }{ \sum_{k \in \mathcal{N}_s^{p_i}} \left( \text{exp} \left( \text{LeakyRELU} \left( \mathbf{h}^{(PS)}_{c_{p_i}} \cdot \mathbf{h}^{(PS)}_{s_k} \right) \right) \right) }}
    \end{gathered}
\end{equation}
where $\mathcal{N}_s^{p_i}$ is neighbour sentences set of $p_i$ in $G_{PS}$. The message aggregation from a neighbour sentence $s_j$ to $p_i$ is computed as: 
\begin{equation}
    \label{mess-agg}
    \mathbf{m}_{p_i \leftarrow s_j}^{(PS)} = \mathbf{\alpha}^{(PS)}_{ij} \mathbf{s}_j
\end{equation}
Finally, the messages from all neighbour sentences of $p_i$ in $G_{PS}$ are accumulated and averaged to obtain new node representation of $p_i$, denoted as $\mathbf{p}^{(PS)}_i$:
\begin{equation}
    \label{sum-repre}
    \resizebox{0.7\hsize}{!}{
        $\mathbf{p}^{(PS)}_i = \frac{1}{\text{H} \times |\mathcal{N}_s^{p_i}|} \left[ \sum_{h=1}^H \left( \sum_{s_k \in \mathcal{N}_s^{p_i}} \mathbf{m}_{p_i \leftarrow s_k}^{(PS)} \right) + \mathbf{b}_p^{(PS)} \right]$}
\end{equation}
where $\text{H}$ is the number of heads in the multi-head attention mechanism; and $\mathbf{b}_p^{(PS)}$ is the bias parameter. 

To derive the representation of sentence $s_j$ within $G_{PS}$, we use Eq (\ref{mess-agg}) to calculate $\mathbf{m}_{s_j \leftarrow p_i}^{(PS)}$ and apply Eq (\ref{sum-repre}) to the set of neighboring pair nodes $\mathcal{N}_p^{s_j}$ of $s_i$.

\paragraph{In $G_{PP}$.} An identical procedure is also implemented for $G_{PP}$ to derive the representation of a pair $p_i$. Concretely, we substitute $\mathbf{c}_{p_i}$ and $\mathbf{s}_j$ with the relational information of $p_i$ and its neighboring pair $p_j$ (i.e., $\mathbf{r}_i$ and $\mathbf{r}_j$) in Eq (\ref{attn-mecha}). Subsequently, we compute $\mathbf{m}_{p_i \leftarrow p_j}^{(PP)}$ and $\mathbf{p}^{(PP)}_i$ by following Eq (\ref{mess-agg}) and Eq (\ref{sum-repre}) respectively.

\subsubsection{Final Node Representation}
We stack $L$ layers of EP-GAT to obtain node representations for entity pairs and sentences for each sub-graph. Specifically, the node representation of sentence $s_j$ in the $l^{th}$ layer, denoted as $\mathbf{s}_j^l$, is derived through the following procedure:
\begin{equation}
    \label{node-l-layer}
    \mathbf{s}_j^l = \frac{1}{|\mathcal{S}^l_{s_j}|} \left( \sum_{\mathbf{s} \in \mathcal{S}^l_{s_j}} \text{LeakyRELU}(\mathbf{s}) \right)
\end{equation}
where $\mathcal{S}^l_{s_j} = \left\{ \mathbf{s}_j^{l-1}, {\mathbf{s}_j^l}^{(PS)}, {\mathbf{s}_j^l}^{(SS)} \right\}$ represents the set of node representations for sentence $s_j$ from the ${(l-1)}^{th}$ layer, $G_{PS}$, and $G_{SS}$ respectively.

Finally, the node representation of sentence $s_j$ after $L$ layers of EP-GAT is as follows:
\begin{equation}
    \mathbf{s}_j = \frac{1}{L} \left( \sum_{l=1}^L \mathbf{s}_j^l \right)
\end{equation}

An analogous process is applied to obtain the final node representation after $L$ graph layers for an entity pair $p_i$. The main difference lies in the formulation of $\mathcal{P}^l_{p_i}$ as stated in Eq (\ref{node-l-layer}) where $\mathcal{P}^l_{p_i} = \left\{ \mathbf{p}_i^{l-1}, {\mathbf{p}_i^l}^{(PS)}, {\mathbf{p}_i^l}^{(PP)} \right\}$.

\subsection{Evidence Prediction}
After applying $L$ layers of EP-GAT, we employ a bilinear function to identify whether a sentence $s_j$ is an evidence sentence of the pair $p_i$ as follows:
\begin{equation}
    \mathbb{P}\left( s_j | p_i \right) = \sigma \left( \mathbf{s}_j^T \mathbf{W}_{er} \mathbf{p}_i + \mathbf{b}_{er} \right)
\end{equation}
where $\mathbb{P}\left( s_j | p_i \right)$ is the probability of sentence $s_j$ being an evidence sentence of entity pair $p_i$; $\mathbf{s}_j$ and $\mathbf{p}_j$ are representation of $s_j$ and $p_i$; $\mathbf{W}_{er}$ and $\mathbf{b}_{er}$ are trainable parameters; and $\sigma$ is the sigmoid function.

\subsection{Training Objective}
We observe there exists a noticeable imbalance in the ratio of positive to negative supporting evidence sentences for a positive entity pair. Specifically, around 95\% of positive entity pairs have a maximum of 3 supporting evidence sentences, with 94\% of these pairs having a maximum of 2 sentences. Meanwhile, the average document length is 8 sentences. Therefore, to mitigate the negative impact of this phenomenon, we utilize the Focal Loss \cite{lin2017focal} as our objective function:
\begin{equation}
    \small
    \label{focal-loss}
    \begin{split}
        \mathcal{L}_{ER} = - \sum_{p\in \mathcal{P}_D} \sum_{s \in \mathcal{S}_D} [\alpha \left( 1 - \mathbb{P}(s|p) \right)^\gamma \text{log}\left( \mathbb{P}(s|p) \right) + \\ (1-\alpha) \mathbb{P}(s|p)^\gamma \text{log}\left( 1 - \mathbb{P}(s|p) \right)]
    \end{split}
\end{equation}
where $\mathcal{P}_D$ and $\mathcal{S}_D$ are set of positive entity pairs and sentences in the document $D$; $\mathbb{P}(s|p)$ is the probability of sentence $s$ is an evidence sentence of positive entity pair $p$; and $\alpha, \gamma$ are hyper-parameters.

\begin{table*}[t]
    \centering
    \caption{Results (\%) on the development and test set of DocRED \cite{yao2019docred}. Experiments are run with 5 different random seeds on the development set. Mean and standard deviation of performance are reported. The result with \textdagger \ is based on our reproduction. Other results are obtained from \cite{xie2022eider}. Bold indicates the best performance.}
    \begin{adjustbox}{max width=\textwidth}
        \begin{tabular}{lcccccccc}
            \hline
            \textbf{Model} & \multicolumn{4}{c}{\textbf{Dev}} &  & \multicolumn{3}{c}{\textbf{Test}} \\
             & \textit{PLM} & \textit{Ign $\text{F}_1$} & \textit{$\text{F}_1$} & \textit{Evi $\text{F}_1$} &  & \textit{Ign $\text{F}_1$} & \textit{$\text{F}_1$} & \textit{Evi $\text{F}_1$} \\ \hline \hline
             BiLSTM \cite{yao2019docred}                                    & - & 48.87 & 50.97 & 44.07 &  & 48.78 & 51.06 & 43.83 \\
             LSR \cite{nan2020reasoning}                                    & $\text{BERT}_{\text{base}}$ & 52.43 & 59.00 & - &  & 56.97 & 59.05 & - \\
             GAIN \cite{zeng2020double}                                     & $\text{BERT}_{\text{base}}$ & 59.14 & 61.22 & - &  & 59.00 & 61.24 & - \\
             E2GRE \cite{huang2021entity}                                   & $\text{BERT}_{\text{base}}$ & 55.22 & 58.72 & 47.14 &  & - & - & 48.35 \\
             Reconstruct \cite{xu2021document}                              & $\text{BERT}_{\text{base}}$ & 58.13 & 60.18 & - &  & 57.12 & 59.45 & - \\
             ATLOP \cite{zhou2021document}                                  & $\text{BERT}_{\text{base}}$ & 59.22 & 61.09 & - &  & 59.31 & 61.30 & - \\
             DocuNet \cite{zhang2021document}                               & $\text{BERT}_{\text{base}}$ & 59.86 & 61.83 & - &  & 59.93 & 61.86 & - \\
             \hline
             EIDER \cite{xie2022eider}\textsuperscript{\textdagger}         & $\text{BERT}_{\text{base}}$ & 60.05 & 62.09 & 49.92 &  & 59.62 & 61.92 & 49.99 \\
             \ \ + \textbf{CDER (Ours)}                                     & $\text{BERT}_{\text{base}}$ & \textbf{60.31\scriptsize$\pm$0.10} & \textbf{62.32\scriptsize$\pm$0.07} & \textbf{52.92\scriptsize$\pm$0.11} &  & \textbf{59.98} & \textbf{62.33} & \textbf{52.89} \vspace{1mm} \\
             \hdashline
             DREEAM \cite{ma2023dreeam}\textsuperscript{\textdagger}        & $\text{BERT}_{\text{base}}$ & 60.35 & 62.35 & 52.78 &  & 60.13 & 62.34 & 52.93 \\
             \ \ + \textbf{CDER (Ours)}                                     & $\text{BERT}_{\text{base}}$ & \textbf{60.42\scriptsize$\pm$0.03} & \textbf{62.42\scriptsize$\pm$0.02} & \textbf{52.86\scriptsize$\pm$0.10} &  & \textbf{60.22} & \textbf{62.45} & \textbf{53.03} \\
             \hline
        \end{tabular}
    \end{adjustbox}
    \label{tab:docre}
\end{table*}

\section{Experiments}
\subsection{Experiment Setup}
\subsubsection{Datasets}
Our ER model is evaluated on the DocRED dataset \cite{yao2019docred}, a large-scale general domain dataset in the DocRE task. The dataset is constructed from Wikipedia and Wikidata, containing a collection of 132,235 entities, 56,354 relational facts, and 96 relation classes. It is the sole DocRE dataset providing evidence sentence label.

\subsubsection{Settings}
Our models are implemented using PyTorch \cite{paszke2019pytorch} and Huggingface's Transformers \cite{wolf2019huggingface} libraries. We utilize the cased BERT$_{\textnormal{base}}$ \cite{devlin2019bert} as document encoder. We employ the AdamW optimizer \cite{loshchilov2017decoupled} with a learning rate of $5e-5$ for the encoder and $1e-4$ for the other parameters. We incorporate linear warmup for the initial 6\% of total steps and set the maximum gradient norm to $1.0$. During training, the batch size, representing the number of documents per batch, is set to 4. However, due to memory constraints as utilizing all possible entity pairs in a document during testing, we set the testing batch size to 1. The threshold $\theta$ is set to $0.5$. Each model undergoes 30 epochs of training on a single NVIDIA Tesla V100 GPU with 32GB of memory. For our objective function in Eq (\ref{focal-loss}), we set the values of $\alpha$ and $\gamma$ to be $0.25$ and $2$, respectively.

We select EIDER \cite{xie2022eider} and DREEAM \cite{ma2023dreeam} as our primary baseline methods for benchmarking ER task. For the DocRE task, our framework acts as a plug-in for any existing DocRE system. Specifically, we retain the DocRE system in baseline methods and substitute their ER results with our framework's output.

\subsubsection{Evaluation Metrics}
We use the \textbf{F1} and \textbf{Ign F1} main evaluation metrics for DocRE task where \textbf{Ign F1} measures the F1 score excluding the relational facts shared by training and dev/test sets \cite{yao2019docred}. For ER, we use the \textbf{PosEvi F1} \cite{xie2022eider} and \textbf{Evi F1} which measure the F1 score of evidence only on positive entity pairs (i.e., those with non-NA relation) and all predicted entity pairs respectively.

\begin{table}[t]
    \centering
    \caption{Results of ER task for positive entity pairs on DocRED dev dataset. Means and standard deviations of 5 runs with different seed numbers are reported. The result with \textdagger \ is based on our reproduction from their released code.}
    \begin{adjustbox}{max width=\textwidth}
        \begin{tabular} {lccc}
            \hline
            \textbf{Method} & \textbf{Pos Evi Precision} & \textbf{Pos Evi Recall} & \textbf{Pos Evi $\mathbf{F_1}$}\\
            \hline
            \hline
            EIDER \cite{xie2022eider}                                       & 77.55 & 78.77 & 78.15 \\
            \hline
            DREEAM \cite{ma2023dreeam}\textsuperscript{\textdagger}         & 86.79 & 82.59 & 84.64 \\
            \hline
            \textbf{CDER (ours)}                                            & \textbf{86.84\scriptsize$\pm$0.31} & \textbf{82.75\scriptsize$\pm$0.35} & \textbf{84.75\scriptsize$\pm$0.05}  \\
            \hline
        \end{tabular}
    \end{adjustbox}
    \label{tab:ee-result}
\end{table}
\vspace{-2mm}
\begin{table}[h]
    \centering
    \caption{ER performance when using all possible entity pairs and only positive entity pairs. Means and standard deviations of 5 runs with different seeds are reported.}
    \begin{adjustbox}{max width=\columnwidth}
        \begin{tabular} {lccc}
            \hline
            \textbf{Method} & \textbf{Pos evi Precision} & \textbf{Pos evi Recall} & \textbf{Pos evi $\mathbf{F_1}$}\\
            \hline
            \hline
            \textbf{CDER} (\textit{positive})   & 86.84\scriptsize$\pm$0.31 & \textbf{82.75\scriptsize$\pm$0.35} & \textbf{84.75\scriptsize$\pm$0.05} \\
            \hline
            \textbf{CDER} (\textit{all})        & \textbf{86.86\scriptsize$\pm$0.2} & 82.63\scriptsize$\pm$0.17 & 84.69\scriptsize$\pm$0.05 \\
            \hline
        \end{tabular}
    \end{adjustbox}
    \vspace{-3mm}
    \label{tab:ee-compare}
\end{table}

\subsection{Results}
\subsubsection{Evidence Retrieval}
We present a comparative examination of the ER performance between our model, CDER, and two existing approaches, EIDER \cite{xie2022eider} and DREEAM \cite{ma2023dreeam}. As reported in Table \ref{tab:ee-result}, CDER surpasses the baseline methods and attains peak performance, exhibiting a notable increase of 6.6 and 0.11 in Pos Evi $\text{F}_1$ score compared to EIDER and DREEAM respectively. Furthermore, our model excels in Pos Evi Precision and Pos Evi Recall metrics, solidifying the efficacy of our approach.

Table \ref{tab:ee-compare} illustrates the experiment configurations on the DocRED development dataset, where all entity pairs (i.e., \textbf{CDER} (\textit{all})) are employed as nodes in both $G_{PS}$ and $G_{PP}$ since positive entity pairs are not known during testing. In contrast to utilizing only positive pairs (i.e., \textbf{CDER} (\textit{positive})), CDER experiences only a marginal decrease of 0.06 in Pos Evi $\text{F}_1$ when utilizing all entity pairs as nodes. This outcome underscores the robustness of our ER approach in a real-world DocRE setting.

We additionally present the ER performance on all predicted entity pairs in Table \ref{tab:docre}, using the $\text{Evi F}_1$ score as the evaluation metric. In comparison to the ER outcomes of EIDER and DREEAM, our method demonstrates a superior $\text{Evi F}_1$ score in both the DocRED development and test datasets, showcasing increases of 3.00 and 0.08, respectively.

\subsubsection{Relation Extraction}
Table \ref{tab:docre} showcases the outcomes of the DocRE task on both the development and test sets of the DocRED dataset. This presentation involves substituting outcomes of our ER model with the one obtained from baseline methods. The table underscores that replacing ER results from CDER can result in increasing $\text{F}_1$ scores and Evi $\text{F}_1$ compared to the baseline methods. Notably, the integration of CDER substantially enhances the performance of the EIDER model, a 0.23 increase in $\text{F}_1$ to closely approximate the DREEAM model. Furthermore, CDER contributes to the advancement of the DREEAM model, an additional 0.07 increase in $\text{F}_1$ to accomplish peak performance.

\begin{table}[t]
    \centering
    \caption{Ablation study evaluated on the DocRED dev dataset.}
    \begin{adjustbox}{max width=\columnwidth}
        \begin{tabular} {lccc}
            \hline
            \textbf{Setting} & \textbf{Precision} & \textbf{Recall} & \textbf{Pos evi $\mathbf{F_1}$}\\
            \hline
            \hline
            \textbf{CDER}                       & \textbf{86.84\scriptsize$\pm$0.31} & 82.75\scriptsize$\pm$0.35 & \textbf{84.75\scriptsize$\pm$0.05} \\
            \hline
            - \textit{w/o} dynamic structure    & 86.56\scriptsize$\pm$0.54 & 82.74\scriptsize$\pm$0.47 & 84.60\scriptsize$\pm$0.12 \\
            \hline
            - \textit{w/o} focal loss           & 85.49\scriptsize$\pm$0.35 & \textbf{83.27\scriptsize$\pm$0.48} & 84.36\scriptsize$\pm$0.20 \\
            \hline
             - \textit{w/o} p2s weight           & 85.91\scriptsize$\pm$0.13 & 81.87\scriptsize$\pm$0.16 & 83.86\scriptsize$\pm$0.11 \\
            \hline
             - \textit{w/o} p2p weight           & 86.39\scriptsize$\pm$0.25 & 81.57\scriptsize$\pm$0.28 & 83.85\scriptsize$\pm$0.11 \\
            \hline
        \end{tabular}
    \end{adjustbox}
    \label{tab:ablation}
\end{table}

\subsection{Ablation Study}
Table \ref{tab:ablation} presents the ablation study to showcase the contribution of each proposed component of the CDER model.

Firstly, removing edge weights in $G_{PS}$ (\textit{w/o p2s weight}) by setting weights to 1 significantly decreased the Pos evi $\text{F}_1$ score, validating the rationale outlined in Section \ref{ep-gat}. A similar impact was observed in $G_{PP}$ (\textit{w/o p2p weight}), emphasizing the advantage of prioritizing relevant entity pairs for superior performance. Secondly, utilizing Binary Cross-Entropy Loss (\textit{w/o focal loss}) resulted in a noticeable reduction in Pos evi $\text{F}_1$, revealing an issue of imbalanced evidence sentences within positive entity pairs. Finally, neglecting the similarity in relational information between entity pairs by using a static structure (i.e., without dynamic removal of edges in $G_{PP}$) led to a marginal 0.15 reduction in the average framework performance, providing additional empirical validation for our findings from Section \ref{graph}.

\section{Conclusion}
In summary, we introduce CDER, an innovative attentional graph-based collaborative ER framework aimed at accurately retrieving evidence sentences for individual entity pairs. CDER leverages the collaborative nature of semantically similar entity pairs by dynamically constructing an Entity Pair sub-graph within the document-level bipartite graph at runtime. We further introduce an attentional graph layer called EP-GAT to refine representations of both entity pairs and sentences within the graph. Empirical evaluations on the large-scale DocRED dataset affirm robustness and superior performance of CDER in ER task.

\bibliographystyle{splncs04}
\bibliography{references}

\end{document}